\documentclass{elsart}
\makeatletter
\newif\if@restonecol
\makeatother

\usepackage{algorithmic}
\usepackage[linesnumbered,ruled,vlined]{algorithm2e}
\usepackage{graphicx}
\usepackage{amssymb}
\usepackage{amsmath}
\usepackage{enumerate}
\usepackage{multirow}
\usepackage{booktabs}
\usepackage{threeparttable}
\usepackage{ulem}
\usepackage{rotating}
\usepackage{url}
\usepackage{cite}
\usepackage[switch,pagewise]{lineno}
\usepackage[usenames]{color}
\usepackage{lineno}
\newtheorem{def1}{Definition}
\begin{document}
\begin{frontmatter}
\title{Random forest for label ranking}

\author[Shanghai]{Yangming Zhou} and
\ead{zhou.yangming@yahoo.com}
\author[Shenzhen,Nottingham]{Guoping Qiu\corauthref{cor}}
\corauth[cor]{Corresponding author.}
\ead{guoping.qiu@nottingham.ac.uk}
\address[Shanghai]{Department of Computer Science and Engineering, East China University of Science and Technology, Shanghai 200237, China}
\address[Shenzhen]{College of Information Engineering, Shenzhen University, Shenzhen 518060, China}
\address[Nottingham]{School of Computer Science, University of Nottingham, Nottingham NG72RD,United Kingdom}

\begin{abstract}

Label ranking aims to learn a mapping from instances to rankings over a finite number of predefined labels. Random forest is a powerful and one of the most successful general-purpose machine learning algorithms of modern times. In this paper, we present a powerful random forest label ranking method which uses random decision trees to retrieve nearest neighbors. We have developed a novel two-step rank aggregation strategy to effectively aggregate neighboring rankings discovered by the random forest into a final predicted ranking. Compared with existing methods, the new random forest method has many advantages including its intrinsically scalable tree data structure, highly parallel-able computational architecture and much superior performance. We present extensive experimental results to demonstrate that our new method achieves the highly competitive performance compared with state-of-the-art methods for datasets with complete ranking and datasets with only partial ranking information. 
\end{abstract}

\end{frontmatter}


\section{Introduction}
\label{Sec:Introduction}

Label ranking aims to learn a mapping from instances to rankings over a finite set of predefined labels. It extends the conventional classification and multi-label classification in the sense that it needs to predict a ranking of all class labels instead of only one or several class labels. Both classification and multi-label classification can be considered as a special case of label ranking learning. Specifically, when only the top label is required, label ranking reduces to a classification problem, when a calibrated label is introduced, label ranking is equivalent to a multi-label classification problem \cite{Furnkranz2008}. Due to its generality, label ranking has found in many practical applications such as natural language processing, recommender systems, bioinformatics and  meta-learning \cite{Furnkranz2010}.

In this paper, we propose a new label ranking method based on random forests (LR-RF). Random forests has been widely used to solve a number of machine learning, computer vision and medical image analysis tasks, and has achieved excellent performances \cite{Criminisi2012,Fu2012}. The random forest has many advantages, which make it competitive for solving label ranking problem to existing approaches. Firstly, the tree structure of our LR-RF makes the retrieval of nearest neighbours more efficient than instance-based approaches. Secondly, both the construction and prediction processes of our proposed method can be executed in a parallel way. To turn the random forest into an effective ranking method, we have proposed a ``top label as class'' method to guide the construction of decision tree, and we also developed a novel two-step rank aggregation strategy which effectively aggregate the neighboring rankings into a final predicted ranking. We identify the main contributions of this work as follows.
\begin{itemize}
	\item We explore the usefulness of random forest to solve the label ranking problem (denoted as LR-RF). A ``top label as class'' method is designed for guiding the construction of decision tree, and we also developed a novel two-step rank aggregation strategy which effectively aggregate the neighboring rankings into a final predicted ranking. Experimental results show that our proposed LR-RF method is highly competitive with the state-of-the-art label ranking algorithms.
	\item We extend our proposed LR-RF to the case with only partial ranking information. In this scenario, our proposed LR-RF significantly better than the state-of-the-art algorithms, the advantage is more obvious as the missing probability increases.
\end{itemize}

The paper is organized as follows. In the following section, we briefly review the related work in the literature. Section \ref{Sec:Label Ranking} describes the label ranking problem in a more formal setting and introduces several common distance measures for rankings. In Section \ref{Sec:Random Forest for Label Ranking}, we present our label ranking method based on the random forest model. Section \ref{Sec:Computational Results} is dedicated to an experimental evaluation of the proposed LR-RF based on benchmark datasets. Section \ref{Sec:Discussion and Analysis} conducts several important investigations of the proposed LR-RF algorithm. Finally, Section \ref{Sec:Conclusions and Future Work} concludes the paper.

\section{Related work}
\label{Sec:Related Work}

Due to the practical significance, label ranking has attracted increasing attention in the recent machine learning literature, and a large number of methods have been proposed or adapted for label ranking \cite{Shalev2006,Hullermeier2008,Cheng2009,Cheng2010,Zhou2014a,Schafer2015,Hullermeier2015,DeSa2016,DeSa2017,Aledo2017,Schafer2018}. An overview of label ranking algorithms can be found in \cite{Vembu2011,Zhou2014b}. Existing label ranking methods can be mainly divided into three categories.

One is known as reduction approaches which transform the label ranking problem into several simpler binary classification problems, and then the solutions of these classification problems are combined into a predicted ranking. Label ranking by learning pairwise preferences and by learning utility functions are two widely used schemes in the reduction approaches. For example, ranking by pairwise comparison (RPC) learns binary models for each pair of labels, and the predictions of these binary models are then aggregated into a ranking  \cite{Hullermeier2008}; while constraint classification (CC) and log-linear models for label ranking (LL) seek to learn linear utility functions for each individual label instead of preference relations for each pair of labels \cite{Har2003,Dekel2004}.

The second category is probabilistic approaches which represent label ranking based on statistical models for ranking data, i.e., parametrized probability distributions on the class of all rankings. For example, Cheng et al. have developed instance-based (IB) learning algorithms based on the Mallows (M) and Plackett-Luce (PL) models \cite{Cheng2009,Cheng2010}, and Zhou et al. proposed a label ranking method based on Gaussian mixture models \cite{Zhou2014a}.

Both reduction approaches and probabilistic approaches have shown good performances in the experimental studies, while they also come with some disadvantages. For reduction approaches, theoretical assumptions on the sought ``ranking-valued'' mapping, which may serve as a proper learning bias, may not be easily translated into corresponding assumptions for the classification problems. Moreover, it is often not clear that minimizing the loss function on the binary problems leads to maximizing the performance of the label ranking model in terms of the desired loss function on rankings \cite{Duchi2010}. For probabilistic approaches, their success also do not come for free but at a large cost associated with both memory and time. For example, the instances-based approaches involve costly nearest neighbour search and the aggregation of neighboring rankings is also slow as it requires using complex optimization procedures, such as the approximate expectation maximization in IB-M and the minorization maximization in IB-PL \cite{Cheng2012}. Both IB-M and IB-PL are lazy learners, with almost no cost at training phase but a higher cost at predicting phase. It can be costly or even impossible in the resources-constrained applications.

Besides the reduction approaches and probabilistic approaches, tree-based approaches are also very popular in label ranking. Several label ranking methods based on decision tree were designed for label ranking. For example, Cheng et al., proposed the first adaptation of decision tree algorithm for label ranking, called label ranking tree (LRT)\cite{Cheng2009}. A new version of decision trees for label ranking called entropy-based ranking tree (ERT) and a label ranking forest using this ERT as base learner were developed \cite{DeSa2017}. Recently, a bagging algorithm which takes the weak LRT-based models as base classifiers was proposed. Experimental results show that bagging these weak learners improves not only the LRT algorithm, but also the instances-based algorithms \cite{Aledo2017}. Actually, our proposed random forest for label ranking (LR-RF) in this work falls this category. In the Section \ref{Sec:Computational Results}, we experimentally compare our LR-RF with the state-of-the-art algorithms from these categories.

\section{Label ranking}
\label{Sec:Label Ranking}

Label ranking can be considered as a natural extension of the conventional classification problem. Given an instance $x$ from an instance space $\mathcal{X}$, instead of predicting one or several possible class labels, label ranking tries to associate $x$ with a total order of all class labels. This means that there exists a complete, transitive and asymmetric relation $\succ_x$ on $\mathcal{L}$, where $\lambda_i \succ_x \lambda_j$ shows that $\lambda_i$ precedes $\lambda_j$ in the ranking assigned to $x$.

We can identify a ranking $\succ_x$ with a permutation $\pi_x$ on $\{1,2,\ldots,m\}$ such that $\pi_x(i)=\pi_x(\lambda_i)$ is the position of $\lambda_i$ in the ranking. This permutation encodes the ranking given by
\begin{equation}
    \lambda_{\pi^{-1}_x(1)} \succ_x \lambda_{\pi^{-1}_x(2)}\succ_x \ldots \succ_x \lambda_{\pi^{-1}_x(m)}
\end{equation}
where $\pi^{-1}_x(i)$ is the index of the class label at position $i$ in the ranking. For example, given a label set $\mathcal{L}=\{\lambda_1, \lambda_2, \lambda_3, \lambda_4, \lambda_5\}$, and an observation over these labels $\lambda_4 \succ \lambda_2 \succ \lambda_3 \succ \lambda_5 \succ \lambda_1$, then we can represent this ranking by a permutation $[5~2~3~1~4]$. The set of all permutations of $\{1,2,\ldots,m\}$ is denoted by $\Omega$. By abuse of terminology, we refer to elements $\pi \in \Omega$ as both permutations and rankings.

Like in classification, we do not assume the existence of a deterministic $\mathcal{X} \rightarrow \Omega$ mapping. Instead, every instance is associated with a probability distribution over $\Omega$ \cite{Cheng2009}. It means that, for each $x \in \mathcal{X}$, there exists a probability distribution $\mathcal{P}(\cdot|x)$ such that, for every $\pi \in \Omega$, $\mathcal{P}(\pi|x)$ is the probability that $\pi$ is the ranking associated with $x$. The objective of label ranking is to learn a model in the form of a mapping  $\mathcal{X} \rightarrow \Omega$. Generally, training data consists of a set of instances $\mathcal{T}=\{\langle x_i , \pi_i \rangle\}, i = 1, 2, \ldots, n$, where $x_i$ is the feature vector containing the value of $d$ feature attributes describing instance $i$, and $\pi_i$ is the corresponding target ranking. Ideally, complete rankings are given as training information. However, it is much more important to allow for incomplete ranking information in the form of a partial ranking
\begin{equation}
\lambda_{\pi^{-1}_x(1)} \succ_x \lambda_{\pi^{-1}_x(2)} \succ_x \ldots \succ_x \lambda_{\pi^{-1}_x(m')}
\end{equation}
where $m' < m$ and $\{\pi^{-1}_x(1),\pi^{-1}_x(2),\ldots,\pi^{-1}_x(m')\} \subset \{1,2,\ldots,m\}$. In the above example, it is possible that only partial ranking information is provided for instance $x$, i.e., $\lambda_4 \succ_x \lambda_3 \succ_x \lambda_1$, while no preference information is available for $\lambda_2$ and $\lambda_5$.

To evaluate the predictive performance of a label ranking algorithm, a suitable evaluation function is necessary. Kendall tau distance \cite{Kendall1955} is one of the most widely used distance measures for rankings. It essentially measures the total number of discordant label pairs (label pairs that are ranked in the opposite order in two rankings). Formally,
\begin{equation}\label{Equ:Kendall tau Distance}
\mathcal{D}_K(\pi,\sigma)=\#\{(i,j)|\pi(i)>\pi(j) \wedge \sigma(i)<\sigma(j)\}
\end{equation}
where $1 \leqslant i < j \leqslant m$. Kendall tau distance is an intuitive and easily interpretable performance measure. The time complexity of computing the Kendall tau distance between two rankings is $O(m\log m)$. By normalizing Kendall tau distance to the interval $[-1,1]$, we can obtain Kendall's tau coefficient,
\begin{equation}\label{Equ:Kendall tau Coefficient}
    \tau = 1- \frac{4\mathcal{D}_K(\pi,\sigma)}{m(m-1)}
\end{equation}
which is a well-known correlation measure. Kendall's tau coefficient measures the proportion of the concordant pairs of labels in two rankings. Therefore, this measure can still work with partial rankings, as long as there is at least one pair of labels per instance. When $\tau =1$, it means that the labels in ranking $\pi$ and $\sigma$ are sorted in the same order, while $\tau=-1$ indicates that the labels in these two rankings are sorted in opposite order. In label ranking, performance comparisons among label ranking algorithms are often based on Kendall's tau coefficient. Two alternative distance measures on rankings include the Spearman distance
\begin{equation}\label{Equ:Spearman Distance}
\mathcal{D}_S(\pi,\sigma)=\sum^m_{i=1}(\pi(i)-\sigma(i))^2
\end{equation}
and the Spearman footrule distance
\begin{equation}\label{Equ:Footrule Distance}
\mathcal{D}_F(\pi,\sigma)=\sum^m_{i=1}|\pi(i)-\sigma(i)|
\end{equation}
Both this two kinds of Spearman distances between two rankings can be computed in linear time $O(m)$. Additionally, all the three distance measures can be extended in a natural way to several rankings. For example, the generalized Kendall distance between a complete $\pi$ and a set of rankings $\sigma_1 ,\ldots,\sigma_k$ is given by
\begin{equation}\label{Extension Kendall Distance}
\mathcal{D}_K(\pi, \sigma_1 ,\ldots,\sigma_k) = \sum^k_{i=1}\mathcal{D}_K(\pi,\sigma_i)
\end{equation}

\section{Random forest for label ranking}
\label{Sec:Random Forest for Label Ranking}

\begin{figure}[!htbp]
\begin{center}
\centerline{\includegraphics[width=0.8\textwidth]{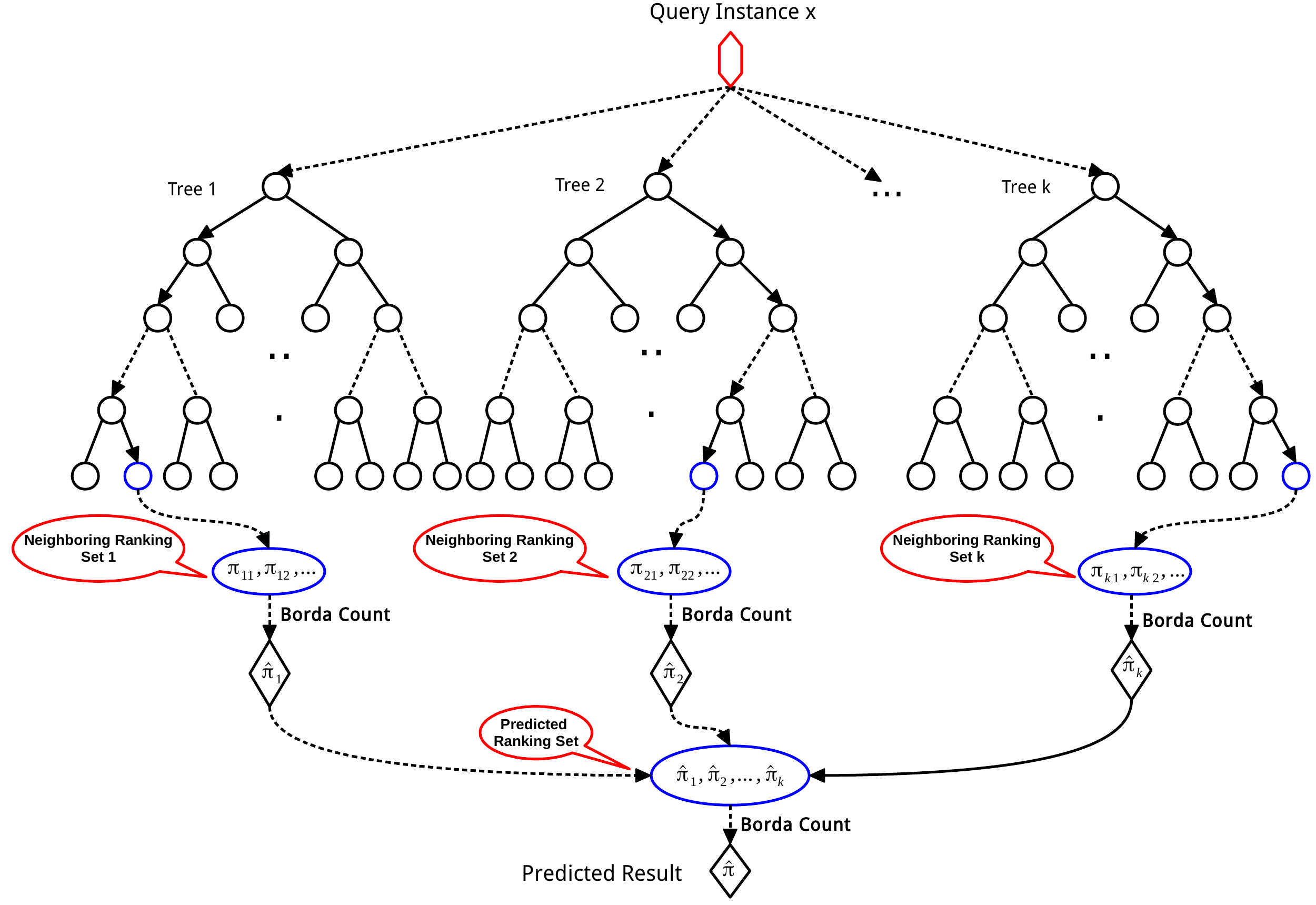}}
\caption{A schematic illustration of random forest for label ranking.}
\label{Fig:LR-RF}
\end{center}
\end{figure}

Random forest is a powerful learning algorithm proposed in \cite{Breiman2001}, which combines several randomized decision trees and aggregates their predictions by averaging. It has been one of the most successful general-purpose algorithms in modern times. In this section, we present a label ranking method based on random forest model, denoted as LR-RF. The proposed LR-RF works in two phases. It first constructs multiple decision trees by using different training instances at construction phase (Section \ref{SubSec:Construction of the Random Forest}), and then at the prediction phase, query instance passes through all trees, a two-step rank aggregation strategy is applied to aggregate the neighboring rankings into a final predicted ranking (Section \ref{SubSec:Prediction Based on Neighboring Rankings}). Fig. \ref{Fig:LR-RF} illustrates the entire process from a query instance $x$ to finally obtain the predicted ranking $\hat{\pi}$

\subsection{Construction of the random forest}
\label{SubSec:Construction of the Random Forest}

To build a random forest, we need to generate $nbr_{tree}$ different training data sets $\mathcal{T}_i ,i=1 \ldots nbr_{tree}$, and each one is used to train a decision tree independently. Each training data set is drawn at random, with replacement, from the original data set $\mathcal{T}$. Then a decision tree is grown on this new training dataset. Specifically, at each node of each tree, a split is performed by maximizing the information gain $\mathcal{G}$ over $n_s = \lfloor \log_2 d \rfloor + 1$  attributes chosen uniformly at random among the $d$ original attributes. Finally, construction of individual tree is stopped until one of the stopping conditions is satisfied. In the forest, all the trees grown are not pruned. To generate such a tree, we partition the training data set $\mathcal{T}$ in a recursive manner, using one-dimensional splits defined by thresholds for an attribute value. The split function at each node is defined as follows
\begin{equation}\label{Equ:Split Function}
\begin{cases}
x^i \geqslant x^i_{threshold} & \text{go to left child}\\
\text{otherwise}, & \text{go to right child}
\end{cases}
\end{equation}
where $x^i$ and $x^i_{threshold}$ are split attribute and split threshold of a best split point, respectively. Based on the above split function, we can split the training data of current node into the left child node and the right child node accordingly. To narrow down the search space for the split function, we only use a small group of attributes to split on rather than using all attributes each time. As different split attributes or split thresholds will bring about different partitions on the training data, it is necessary to find a best split point for each node. In our algorithm, we use the top class labels of the rankings associated with each instance as supervising information to find the best split and guide the growing of the random trees.

In recent years, many methods have been proposed for using ranking information to guide the construction of decision tree. The simplest is Ranking As Class (RAC) \cite{deSa2013}, which simply treats rankings as classes:
$\forall \pi_i \in \Omega, \pi_i \rightarrow \lambda_i$. This allows the use of many supervised construction methods developed for classification in label ranking problems. However, if the RAC approach is applied, the number of classes can be extremely large, up to a maximum number of $m!$, where $m$ is the total number of labels. Additionally, two complicated and specially designed methods have also been proposed for label ranking, such as Minimum Description Length Principle for Ranking data (MDLP-R)\cite{deSa2013} and Entropy-based Discretization for Ranking (EDiRa) \cite{DeSa2016}. Specifically, EDiRa takes into account the properties of rankings: how many distinct rankings are present in current node, and how similar they are to each other.

It is difficult to determine which methods is best to guide the construction of decision tree because many evaluation measures can be used to evaluate their performance \cite{Garcia2013}. In fact, all aforementioned methods can be used for handling ranking information. In our case, we make a compromise between the simple and the complicated methods, and propose to use the top label of a ranking to determine a class, i.e., Top Label As Class (TLAC). TLAC replaces the rankings by classes only consider the top label in the rankings regardless of other labels. For example, in Table \ref{Tab:Iris Example}, even though example 2 and 3 have distinct rankings $[2~3~1]$ and $[3~2~1]$ respectively, TLAC assigns the same class $\lambda_3$ to them because their top labels are same.

\begin{table}[!ht]
\centering
\caption{Some selected examples from iris data set.}
\label{Tab:Iris Example}
\begin{center}
\begin{tabular}{ccccccc}
\toprule[0.75pt]
TID    & $x^1$ & $x^2$ & $x^3$ & $x^4$ & $\pi$ & $\lambda^{TLAC}$ \\
\midrule[0.5pt]
1 &-0.556  &0.250  &-0.864 &-0.917& $[1~2~3]$ & $\lambda_1$\\
2 & 0.167  &0.000  &0.186  &0.167 & $[2~3~1]$ & $\lambda_3$\\
3 & 0.222  &-0.167 &0.424  &0.583 & $[3~2~1]$ & $\lambda_3$\\
4 & 0.056  &0.167  &0.492  &0.833 & $[3~1~2]$ & $\lambda_2$\\
5 & -0.611 &-1.000 &-0.153 &-0.250& $[2~1~3]$ & $\lambda_2$\\
6 & -0.111 &-0.167 &0.085  &0.167 & $[2~3~1]$ & $\lambda_3$\\
\bottomrule[0.75pt]
\end{tabular}
\end{center}
\end{table}

With the help of TLAC, the well-known information gain is become available in label ranking. Information gain is a widely-used splitting criterion to find the best split points in decision trees \cite{Quinlan1986}. It essentially measures the change of class entropy before and after the partition caused by a split point. A split point at each node is represented by a split attribute and corresponding split threshold. For each split point, the entropy of the original data is compared with the weighted sum of the entropy of data in the left node and the right node. Let $\mathcal{G}$ be the information gain, then at the split node $j$, the information gain obtained by a split point $\theta$ in attribute A is defined as
\begin{equation}
	\mathcal{G}(A,\theta;\mathcal{T}_j) = E(\mathcal{T}_j)-\sum_{i \in \{l, r\}} \frac{\vert \mathcal{T}^i_j \vert}{\vert \mathcal{T}_j \vert} E(\mathcal{T}^i_j)
\end{equation}
where $\vert \mathcal{T}_j \vert$ represents the number of instances contained in current node $j$, while $\vert \mathcal{T}^l_j \vert$ and $\vert \mathcal{T}^r_j \vert$ are the number of instances on its left child node and the number of instances on the right child node, respectively, with the split point $\theta$ in attribute $A$. The entropy for data in node $j$ is defined as
\begin{equation}
	E(\mathcal{T}_j)=-\sum_{{\lambda_i}'} p({\lambda_i}') \log p({\lambda_i}')
\end{equation}
where $p({\lambda_i}')$ represents the proportion of instances whose top label of the ranking is ${\lambda_i}'$ in the data set $\mathcal{T}_j$.

A good split means it minimizes the overall entropy in its left child node and right child node. This can be achieved by finding the best split point at node $j$ by
\begin{equation} \label{Equ:Calculate Split Point}
\theta^{*}= \arg \max_{\theta \in \mathcal{S}_j} \mathcal{G} (A,\theta;\mathcal{T}_j)
\end{equation}
where $\mathcal{S}_j$ is the set of all potential split points at node $j$. Based on this split criteria, the decision tree always chooses a split point which effectively splits the data at current node. This can be done recursively until the depth of the tree reaches a maximum allowable depth $d_{max}$ or the entropy of data in current node is less than $\epsilon_0$.

\subsection{Prediction based on neighboring rankings}
\label{SubSec:Prediction Based on Neighboring Rankings}

Once the whole random forest is constructed, it can be used to predict the potential ranking associated with a query instance. During prediction phase, we pass a query instance through all trees simultaneously (starting at the root node) until it reaches the leaf nodes. We call training examples stored in the leaf node as the neighbors of the query sample, and the rankings associated with neighbors as neighboring rankings. The predicted ranking is then obtained by aggregating these neighboring rankings with a two-step rank aggregation procedure.

The problem to aggregate the rankings of neighbors into a ranking is known as \textit{rank aggregation} \cite{Lin2010}. Rank aggregation has been studied extensively in the context of social choice theory and meta-search \cite{Cohen1999}. Rank aggregation can be obtained by optimizing different rank distance measures. For example, when Kendall tau distance is optimized, the achieved rank aggregation is called \textit{Kemeny optimal aggregation}. It has been shown that the Kemeny optimal aggregation is the best compromise ranking. However, finding a Kemeny optimal aggregation is $\mathcal{NP}$-hard even when $k=4$ \cite{Bartholdi1989}. In our algorithm, we resort to an efficient procedure called \textit{Borda's method} to approximately solve it. The Borda's method was originally applied to aggregate label rankings by Klaus Brinker and Eyke H{\"u}llermeier \cite{Brinker2006}. The principle of Borda's method is shown in Definition \ref{Lem:Borda's Method}.

\begin{def1}[Borda's method]\label{Lem:Borda's Method}
Given a collection of complete rankings $\sigma_1 ,\ldots,\sigma_k$, for each label $\lambda_i \in \mathcal{L}$ and ranking $\sigma_j$, Borda's method first assigns a score $s_{ij}=\sigma_j(i)$, and then the average Borda score $s_i$ is defined as $\frac{1}{k}\sum^k_{j=1}s_{ij}$. The labels are then sorted in decreasing order of their average Borda score, and ties are broken at random.
\end{def1}

The rank aggregation obtained by Borda's method is an optimal aggregation with respect to the Spearman distance $\mathcal{D}_S$ \cite{Hullermeier2004}. Moreover, it has been shown that Kendall tau distance $\mathcal{D}_K$ can be approximated very well by Spearman distance $\mathcal{D}_S$ \cite{Diaconis1977}, i.e., $\frac{1}{\sqrt{m}}\mathcal{D}_K(\pi,\sigma) \leqslant \mathcal{D}_S(\pi,\sigma) \leqslant 2\mathcal{D}_K(\pi,\sigma)$, where $m$ is the number of labels in the ranking. It means that there is a close relation between Kendall tau distance and Spearman distance. Therefore, rank aggregation obtained by Borda's method can be a good approximation of the Kemeny optimal aggregation without much sacrifice of predictive performance.

A primary advantage of Borda's method is that it is computationally very easy, and it can perform the aggregation of $k$ complete rankings with $m$ labels in linear time $O(k \cdot m)$. However, it also shows its shortcomings in generalising to partial rankings. To solve this problem, Cheng et al. \cite{Cheng2009} proposed a generalized Borda's method (as shown in Definition \ref{Lem:Generalized Borda's Method}) to extend traditional Borda's method to partial rankings is by apportioning all the excess score equally among all missing candidates.

\begin{def1}[Generalized Borda's method]\label{Lem:Generalized Borda's Method}
Given a set of partial rankings $\sigma_1 ,\ldots, \sigma_k$, for each label $\lambda_i$ and partial ranking of $m' < m$ labels, if it is a missing label, then it receives $s_{ij}=(m+1)/2$ votes; if it is an existing label with rank $r \in \{1,\ldots,m'\}$, then its Borda score is $s_{ij}=(m'+1-r)(m+1)(m'+1)$. The average Borda score $s_i$ is defined as $\frac{1}{k}\sum^k_{j=1}s_{ij}$. The labels are then sorted in decreasing order of their average Borda score.
\end{def1}

In the prediction phase of our LR-RF, we use a two-step rank aggregation strategy to aggregate the neighboring ranking. At the first step, we consider each decision tree individually, and each decision tree makes the same contribution to generate the final predicted ranking. Therefore, we need to aggregate all the rankings of neighbors, and a rank aggregation can be obtained at each tree. At the second step, we aggregate all the predicted ranking of $nbr_{tree}$ trees into a final predicted ranking. In the two-step rank aggregation procedure, we need to perform $nbr_{tree} + 1$ aggregations in total. It is noteworthy that both the construction and prediction of each decision tree can be executed in parallel. The pseudo-code of the proposed LR-RF algorithm can be found in Algorithm \ref{Alg:LR-RF Algorithm}.

\begin{algorithm}[!htb]
\caption{Pseudo-code of the proposed LR-RF Algorithm}
\label{Alg:LR-RF Algorithm}
\KwIn{a given query instance $x$, training data $\mathcal{T}$, number of trees $nbr_{tree}$ and tree depth $d_{max}$.}
\KwOut{a predicted ranking $\pi_x$ for the query instance $x$}
\Begin{
/$*$ \textbf{Construction phase} $*$/ \\
\For{$i = 1,2,\ldots,nbr_{tree}$}{
	create a new data set $\mathcal{T}_i$ of size $n$ from $\mathcal{T}$;\\
	\While{a stopping condition is not met}{
		select $n_s$ attributes from the original $d$ attributes;\\
   		find the split attribute A and its best split point $\theta^*$;\\
   		split the data $\mathcal{T}_j$ according to the $\theta^*$ in attribute A;\\
	}
}
/$*$ \textbf{Prediction phase} $*$/ \\
\For{$i = 1,2,\ldots,nbr_{tree}$}{
   	pass the query $x$ through $i$-th tree;\\
   	find the nearest neighbors of $x$ in the tree;\\
   	aggregate all the neghboring rankings into a predicted ranking $\pi_i'$ by Borda's method (first-step);\\
}
aggregate $nbr_{tree}$ rankings $\pi_1' ,\ldots, \pi_{nbr_{tree}}'$ into a final predicted ranking $\pi_x$ by Borda's method (second-step);
}
\Return a predicted ranking $\pi_x$
\end{algorithm}

Recently, a label ranking forest (LRF) was proposed in literature \cite{DeSa2017}. LRF adopts the random forest framework as well as our proposed LR-RF. Both LRF algorithm and our proposed LR-RF algorithm adopt the framework of random forest. However, our LR-RF distinguishes itself from LRF algorithms by following four aspects. (1) Entropy-based discretization method is used to construct decision trees in LRF, while our LR-RF employs a very simple method (i.e., Top Label As Class) to guiding the construction of decision tree (see Section \ref{SubSec:Construction of the Random Forest}). (2) Our LR-RF is significantly better than LRF in terms of computational accuracy (see Section \ref{SubSec:Compared With Reported Results of the State-of-the-art Algorithms}). (3) LRF was only evaluated on 15 datasets with complete ranking information, while our LR-RF performs well on both 16 datasets with complete ranking information and 16 datasets with incomplete ranking information (see Section \ref{Sec:Computational Results}).

\subsection{Complexity analysis}
\label{SubSec:Complexity Analysis}

In this section, we discuss the computational complexity of the proposed LR-RF method and the state-of-the-art methods. We consider the case of dataset with complete ranking information.

We suppose that the dataset is composed of $n$ instances, and each instance has $d$ attributes and is associated with a ranking of $m$ class labels. Our proposed LR-RF algorithm has two key components: decision tree and Borda's method. Borda's method is used to obtain the consensus ranking by aggregating the rankings. Its computational complexity is $O(m \cdot n + m \log (m))$, where the first part is due to counting and the second part is for the sorting process \cite{Aledo2017}. For each decision tree, we suppose that the maximum depth of the tree is $d_{max}$. At each level of the tree, each attribute is processed once over the whole dataset. The complexity of the learning process is $d_{max}$ multiplied by the complexity of the operations performed at each level. Based on the TLAC strategy, the decision tree stops when all rankings have same top labels or it reaches the maximum allowable tree depth $d_{max}$. At each node, $n_s = \lfloor \log_2 d \rfloor + 1$ attributes are randomly selected from $d$ attributes to determine the best one. For each attribute $x_i$, a good split point $\theta$ is identified according to Eq. (\ref{Equ:Calculate Split Point}) of complexity $O(m^2 \cdot n)$. We calculate the spilt point for each one of $n_s$ attributes selected from $d$ attribute and considering every value $n$ as threshold. Therefore, the overall complexity for constructing such a decision tree is $O(n_s \cdot m^2 \cdot n^2 \cdot d_{max}$).

Our proposed LR-RF algorithm is composed of $nbr_{tree}$ decision trees. At the training phase, the total computational complexity of our proposed LR-RF algorithm is $O(nbr_{tree} \cdot n_s \cdot m^2 \cdot n^2 \cdot d_{max})$. While for the predicted phase, the complexity of our proposed LR-RF algorithm is $((m \cdot n + nbr_{tree} \cdot m \log (m)) +(m \cdot nbr_{tree} + m \log (m)))$ where the first part is for generating $nbr_{tree}$ predicted ranking by each tree, and the second part is for aggregating $nbr_{tree}$ rankings into a final predicted ranking.

Label ranking tree (LRT) was originally proposed in \cite{Cheng2009}. The computational complexity for the whole learning process of LRT is $O(d \cdot m^2 \cdot n^2 \cdot \log(n))$, as indicated in \cite{Aledo2017}. To reduce the complexity of computing the splitting point, Aledo et al., presented two modified versions of the LRT algorithms in \cite{Aledo2017}. Specially, the splitting points were selected by using two well-known unsupervised discretization criteria: equal-width and equal-frequency. Therefore, the complexity of the whole learning process is reduce to $O(d \cdot m^2 \cdot n \cdot \log(n))$. That is, the complexity regarding LRT has been reduced by $O(n)$. Compare to the original LRT, our decision tree is more efficient, i.e., $n_s = \lfloor \log_2 d \rfloor + 1 \leqslant d$. However, two modified versions of the LRT algorithm are more efficient than the decision tree used in our LR-RF.

Let $nbr_{pr}$ be the number of pairwise preferences that are associated with instance $x_i$, and $z = \frac{1}{n}\sum^n_{i = 1}|nbr_{pr}| \leqslant \frac{m(m-1)}{2}$ be the average number of pairwise preferences over all instances. Given a base learner with complexity $O(n^c)$, the complexity of label ranking by learning pairwise preferences (RPC for short) is $O(zn^c)$ and the complexity of constraint classification (CC for short) is $O(z^c \cdot n^c)$. For the cases that a base learner with a polynomial time complexity ($i.e., c \geqslant 1$), RPC is at least as efficient as CC. Otherwise CC is faster. The total complexity of the boosting-based algorithm proposed for log-linear models (LL for short) is $O(z \cdot n+d \cdot m) \cdot nbr_{iter}$, where $d$ is the number of attributes of an instance and $nbr_{iter}$ is the number of iterations \cite{Hullermeier2008}. For the complexity of the instance-based methods with Mallows model (IB-M for short) and with the Plactett-Luce model (IB-PL for short), a direct comparison is complicated because IB-M and IB-PL are lazy learners, with almost no cost training phase but high cost at prediction phase. Although they have complex local estimation procedures, their implementations are very efficient and quite comparable to the corresponding counterpart for classification \cite{Cheng2012}.

\section{Computational results}
\label{Sec:Computational Results}

In this section, we present an empirical evaluation of the proposed LR-RF with main state-of-the-art label ranking methods.

\subsection{Data sets}
\label{SubSec:Data Sets}

In our experiments, we evaluate the proposed LR-RF on the label ranking data sets from the KEBI Data Repository. These data sets are obtained by transforming multi-class and regression data sets from the UCI repository of machine learning databases and the Statlog collection into label ranking data sets in two different ways. (A) For classification data, a naive Bayes classifier is first trained on the complete data set. Afterwards, for each instance, all the class labels in the data set are ordered according to their predicted class label probabilities, breaking ties by ranking the class label with lower index first. (B) For regression data, some numerical attributes are removed from the set of predictors, and each one is treated as a label. To obtain a ranking, the attributes are standardized and then ordered by size. A summary of the benchmark data sets and their characteristics is provided in Table \ref{Tab:Data Sets}\footnote{All label ranking data sets are publicly available at website: http://www.uni-marburg.de/fb12/kebi/research/repository/labelrankingdata}.

\begin{table}[!ht]
\centering
\caption{The description of experimental data sets (the type indicates the way in which the data set has been generated).}
\label{Tab:Data Sets}
\begin{center}
\begin{tabular}{lrrrr}
\toprule[0.75pt]
Data sets    & type & \#instances & \#features & \#labels \\
\midrule[0.5pt]
authorship   & A & 841   & 70 & 4  \\
bodyfat      & B & 452   & 7  & 7  \\
calhousing   & B & 37152 & 4  & 4  \\
cpu-small    & B & 14744 & 6  & 5  \\
elevators    & B & 29871 & 9  & 9  \\
fried        & B & 73376 & 9  & 5  \\
glass        & A & 214   & 9  & 6  \\
housing      & B & 906   & 6  & 6  \\
iris         & A & 150   & 4  & 3  \\
pendigits    & A & 10992 & 16 & 10 \\
segment      & A & 2310  & 18 & 7  \\
stock        & B & 1710  & 5  & 5  \\
vehicle      & A & 846   & 18 & 4  \\
vowel        & A & 528   & 10 & 11 \\
wine         & A & 178   & 13 & 3  \\
wisconsin    & B & 346   & 16 & 16 \\
\bottomrule[0.75pt]
\end{tabular}
\end{center}
\end{table}

\subsection{Experimental settings}
\label{SubSec:Experimental Settings}

Results were obtained in terms of Kendall's tau coefficient from five repetitions of a ten-fold cross-validation. At each repetition, the dataset is randomly partitioned into 10 equal parts (or folds). Of the 10 parts, a part is retained as the validation data for testing the algorithm, and the remaining 9 parts are used as training data. The cross-validation process is then repeated 10 times, with each part uses exactly once as the validation data. The 10 results from the folds can then be averaged to yield an overall result. To model incomplete observations, we modified the training data according to the following rule. Given a ranking,  we associate each label with a random probability. If the probability is less than a fixed missing probability $p_0$ $(0 \leqslant p_0 \leqslant 1)$, we delete it from the ranking and keep it in the ranking otherwise. Hence, $p_0 \times 100\%$ of the labels in the original data sets will be deleted on average. It is a general practice to model incomplete instances in label ranking \cite{Cheng2009,Cheng2010}.

Our proposed LR-RF\footnote{The source code of the proposed LR-RF algorithm is now publicly available at GitHub: https://yangmingzhou.github.io/LabelRanker/.} algorithm was implemented by MATLAB 2014b running on a MacBook Pro with an Intel Core i5 processor (2.6 GHz and 8 GB RAM). To run our LR-RF algorithm,  there are three parameter values needed to decide in advance. We set the number of decision trees $nbr_{tree} = 50$ in our random forest, the maximum allowable depth of the decision tree $d_{max} = 8$. These two parameter values are determined based on the experimental analysis and discussion in Section \ref{SubSec:Sensitivity Analysis on the Number of Trees} and \ref{SubSec:Sensitivity Analysis on Tree Depth} respectively.

\subsection{Experimental results}
\label{SubSec:Experimental Results}

In recent years, a variety of label ranking algorithms have been proposed in the literature \cite{Vembu2011,Zhou2014b}. The representative algorithms mainly include constraint classification (CC) \cite{Har2003}, log-linear model (LL) \cite{Dekel2004}, ranking by pairwise comparison (RPC) \cite{Hullermeier2008}, instance-based methods with the Mallows model (IB-M) \cite{Cheng2009} and the Plackett-Luce model (IB-PL) \cite{Cheng2010}, label ranking tree (LRT) \cite{Cheng2009} and label ranking forest (LRF) \cite{DeSa2017}. However, some source codes or executed programs of these algorithms are not available. We are only able to compare the performance of the proposed LR-RF algorithm with the reference algorithms RPC, IB-PL, and LRT by means of the WEKA-LR\footnote{The package WEKA-LR is publicly available at: https://www.uni-marburg.de/fb12/kebi/research/software}. WEKA-LR is a label ranking extension for WEKA. It implements three label ranking algorithms, including RPC, IB-PL, and LRT, only in case of complete ranking information. For incomplete ranking information, it is still not available. In the following, we focus on comparing the proposed LR-RF algorithm with these three algorithms. With the help of WEKA-LR, we run these algorithms with default parameters in our platform. All the experimental results are obtained based on 10-fold cross-validation.
\begin{itemize}
	\item Ranking by pairwise comparisons (RPC) has two default parameters. Specifically, the logistic regression is selected as the base classifier, and the voting scheme is soft voting \cite{Hullermeier2008}.
	\item Instance-based methods with the Plackett-Luce model (IB-PL) has one default parameter, i.e., the number of  nearest neighbours $k$ used for prediction of new rankings. In the experiments, we run IB-PL with four kinds of different neighborhood size (i.e., $\{5,10,15,20\}$) \cite{Cheng2010}.
	\item Label ranking tree (LRT) does not need to set any parameter \cite{Cheng2009}.
\end{itemize}

To analyse these results, we use a two-step statistical test procedure to perform performance comparisons \cite{Demsar2006}. Firstly, we conduct a \textsl{Friedman test} which makes the null hypothesis that all algorithms are equivalent. If the null hypothesis is rejected, we then proceed with a post-hoc test named \textsl{two-tailed Bonferroni-Dunn test}. Both Friedman test and the two-tailed Bonferroni-Dunn test are based on the average ranks. We order the algorithms for each data set separately, the best performing algorithm obtaining the rank of $1$, the second best rank $2$, and so on. In case of ties, average ranks are assigned. Finally, we obtain the average rank of each algorithm by averaging the ranks of all 16 datasets.

%
%

\begin{table}[!ht]
\caption{Comparative results between the proposed LR-RF algorithm with state-of-the-art algorithms in case of complete ranking. For each dataset, the rank of each algorithm is indicated in parentheses. The number of winning datasets of each algorithm is also provided at the last row.}
\label{Tab:Comparison Results on Complete Ranking}
\begin{center}
\begin{scriptsize}
\begin{threeparttable}
\begin{tabular}{c|ccccccc}
\toprule[0.75pt]
Data sets & RPC & IB-PL$_5$ & IB-PL$_{10}$ & IB-PL$_{15}$ & IB-PL$_{20}$ & LRT & \textbf{LR-RF} \\
\midrule[0.5pt]
authorship   &0.908(2.0)&-0.203(4.0)&-0.271(5.0)&-0.276(6.0)&-0.344(7.0)&0.887(3.0)&\textbf{0.913}(1.0)\\
bodyfat      &0.282(2.0)& \textbf{0.410}(1.0)& 0.095(6.0)& 0.170(4.0)& 0.070(7.0)&0.110(5.0)&0.185(3.0)\\
calhousing   &0.244(7.0)& 0.341(5.0)& 0.295(6.0)& 0.401(2.0)& \textbf{0.471}(1.0)&0.357(4.0)&0.367(3.0)\\
cpu-small    &0.449(4.0)& 0.400(6.0)& \textbf{0.515}(1.5)& 0.490(3.0)& 0.367(7.0)&0.423(5.0)&\textbf{0.515}(1.5)\\
elevators    &0.749(3.0)& 0.721(5.5)& 0.721(5.5)& 0.721(5.5)& 0.721(5.5)&\textbf{0.756}(1.5)&\textbf{0.756}(1.5)\\
fried        &\textbf{0.999}(1.0)& 0.760(4.0)& 0.508(5.0)& 0.421(6.0)& 0.394(7.0)&0.890(3.0)&0.926(2.0)\\
glass        &0.885(7.0)& \textbf{1.000}(2.5)& \textbf{1.000}(2.5)& \textbf{1.000}(2.5)& \textbf{1.000}(2.5)&0.889(5.0)&0.888(6.0)\\
housing      &0.670(3.0)& 0.266(4.0)&-0.048(7.0)&-0.018(5.0)&-0.032(6.0)&\textbf{0.803}(1.0)&0.792(2.0)\\
iris         &0.889(5.0)& \textbf{0.978}(1.5)& \textbf{0.978}(1.5)& 0.858(6.0)& 0.804(7.0)&0.960(4.0)&0.966(3.0)\\
pendigits    &0.932(4.0)& \textbf{0.975}(1.0)& 0.840(5.0)& 0.812(6.0)& 0.787(7.0)&0.943(2.0)&0.939(3.0)\\
segment      &0.934(7.0)& \textbf{1.000}(2.5)& \textbf{1.000}(2.5)& \textbf{1.000}(2.5)& \textbf{1.000}(2.5)&0.953(6.0)&0.961(5.0)\\
stock        &0.779(6.0)& 0.904(2.0)& 0.851(4.0)& 0.808(5.0)& 0.737(7.0)&0.889(3.0)&\textbf{0.922}(1.0)\\
vehicle      &0.850(6.0)& 0.929(3.0)& \textbf{0.940}(1.0)& 0.930(2.0)& 0.857(5.0)&0.831(7.0)&0.860(4.0)\\
vowel        &0.652(4.0)& 0.841(2.0)& 0.451(5.0)& 0.279(6.0)& 0.162(7.0)&0.794(3.0)&\textbf{0.967}(1.0)\\
wine         &0.903(6.0)& 0.940(5.0)& \textbf{1.000}(1.0)& 0.989(3.0)& 0.996(2.0)&0.884(7.0)&0.953(4.0)\\
wisconsin    &\textbf{0.633}(1.0)& 0.612(2.0)& 0.063(5.0)&-0.028(7.0)& 0.049(6.0)&0.351(4.0)&0.478(3.0)\\
\midrule[0.5pt]
avg.rank	 & 4.25 & 3.19 & 3.97 & 4.47 & 5.41 & 3.97 & \textbf{2.75} \\
\bottomrule[0.75pt]
\end{tabular}
\begin{tablenotes}
\tiny
     \item IB-PL$_x$ means that IB-PL algorithm neighborhood size $x$.
\end{tablenotes}
\end{threeparttable}
\end{scriptsize}
\end{center}
\end{table}

Table \ref{Tab:Comparison Results on Complete Ranking} represents the comparative results between the proposed LR-RF algorithm and the state-of-the-art algorithms on data sets with complete ranking information. The best performance on each data set is in bold. We clearly observe that there is no a single algorithm can achieve the best performance across all 16 data sets. Compared to other reference algorithms, our proposed LR-RF achieves highly competitive performance. Specifically, the proposed LR-RF algorithm achieves the best performance on 5 out of 16 data sets, and obtaining the smallest average ranks 2.75 (4.25, 3.97, 3.19, 3.97, 4.47 and 5.41 are respectively obtained by the reference algorithm RPC, LRT and IB-PL with neighborhood sizes 5, 10, 15, and 20).

We resort to two-step statistical method to check the significant difference among these algorithms. According to the Friedman test, we calculate $F_F = 2.93$. With seven algorithms and 16 data sets, $F_F$ is distributed according to the $F$ distribution with $7-1 = 3$ and $(7-1) \times (16-1) = 90$ degrees of freedom. The critical value of $F(6,90)$ at the significance level 0.05 is 2.21 ($2.93 > 2.21$),so we reject the null-hypothesis. We then proceed with a post-hoc two-tailed Bonferroni-Dunn test in order to compare our LR-RF algorithm with other reference algorithms. At the same significance level, the Critical Difference (CD) for the two-tailed Bonferroni-Dunn in our experiment is $2.638 \sqrt{\frac{7 \cdot 8}{6 \cdot 16}} = 2.01$. We observe that our LR-RF is only significantly better than reference algorithm IB-PL with neighborhood size 20 ($5.41 - 2.75 = 2.66 > 2.01$). Compared with remaining five reference algorithms, our LR-RF is better but there is no significant difference.

\subsection{Compared with reported results of the state-of-the-art algorithms}
\label{SubSec:Compared With Reported Results of the State-of-the-art Algorithms}

Given that source codes or executable programs of reference algorithms are not available in case of partial ranking information. In this section, we compare the performance of the proposed LR-RF algorithm with results reported in the literature. By comparing the results reported and the results obtained by running WEKA-LR, we obverse there are only some slight differences for RPC, IB-PL and LRT. Moreover, it is well known that the executed platform is only closely related to computational time of an algorithm. Consequently, it is also credible to compare the results (in terms of accuracy) of the proposed algorithm with results reported of the reference algorithms. The reference algorithms mainly include constraint classification (CC) \cite{Har2003}, log-linear model (LL) \cite{Dekel2004}, ranking by pairwise comparison (RPC) \cite{Hullermeier2008}, instance-based methods with the Mallows model (IB-M) \cite{Cheng2009} and Plactett-Luce model (IB-PL) \cite{Cheng2010}, label ranking tree (LRT) \cite{Cheng2009}, three ensemble approaches constructed by respectively applying bagging to the three LRT based algorithms (LRT+bag, W-LRT+bag and F-LRT+bag) \cite{Aledo2017}, and two label ranking forests based on ranking trees (LRF-RT for short) and entropy-based ranking trees (LRF-ERT) respectively. It is worth noting that the results of all reference algorithms are taken directly from the literature \cite{Cheng2009, Cheng2012,Aledo2017}, and their experimental settings are displayed as follows.
\begin{itemize}
	\item Constraint classification (CC) is an online-variant as proposed in \cite{Har2003}, using a noise-tolerant perceptron algorithm as a base learner.
	\item Log-linear models for label ranking (LL) uses the linear combination base ranking functions, which map instance/label pairs to real numbers \cite{Dekel2004}.
	\item Ranking by pairwise comparison (RPC) uses the logistic regression as the base learner \cite{Hullermeier2008}.
	\item Label ranking tree (LRT) does not need to set any parameter \cite{Cheng2009}.
	\item For the instance-based methods with the Mallows model (IB-M) and Plackett-Luce model (IB-PL), the neighborhood size $k \in \{5,10,15,20\}$ is determined through cross validation on the training set. To guarantee a fair comparison, we used the Euclidean distance as a distance metric on the instance space \cite{Cheng2009,Cheng2010,Cheng2012}.
	\item The ensemble approaches constructed by applying bagging to the three LRT based algorithms (LRT, W-LRT and F-LRT): LRT+bag, W-LRT+bag and F-LRT+bag, where W-LRT and F-LRT are two modification versions of the LRT algorithm by respectively using unsupervised discretization criteria: equal width and equal frequency \cite{Aledo2017}.
    \item Label ranking forest (LRF) was proposed in literature \cite{DeSa2017}, which builds a label ranking forest based on ranking trees (LRF-RT for short) or entropy-based ranking trees (LRF-ERT) \cite{DeSa2017}. Since the results of LRF is not available for the case of incomplete ranking information, we only consider the results of LRF in complete ranking information (see Table \ref{Tab:Comparison Results P = 0.0}).
\end{itemize}

Table \ref{Tab:Comparison Results P = 0.0}-\ref{Tab:Comparison Results P = 0.6} show performance comparisons between the proposed LR-RF algorithm and state-of-the-art label ranking algorithms on three cases: complete ranking ($p_0 = 0.0$), missing labels ranking with $p_0 = 0.3$ and missing labels ranking with $p_0 = 0.6$. All these results are obtained by five times 10 folds cross-validation. Compared to state-of-the-art algorithms, we clearly observe that our proposed LR-RF achieves the best performance in terms of average rank in all three cases. Moreover, when the missing probability $p$ increases, the average rank of our LR-RF decreases.

\begin{sidewaystable}
\caption{Comparative results between LR-RF with state-of-the-art algorithms on the case $p_0 = 0.0$.}
\label{Tab:Comparison Results P = 0.0}
\begin{center}
\begin{tiny}
\begin{tabular}{l|cccccccccccccc}
\toprule[0.75pt]
\multicolumn{1}{c}{} & \multicolumn{3}{c}{Reduction methods} & \multicolumn{1}{c}{} & \multicolumn{2}{c}{Probabilistic methods} & \multicolumn{1}{c}{} & \multicolumn{7}{c}{Tree-based methods}\\
\cmidrule[0.5pt]{2-4} \cmidrule[0.5pt]{6-7} \cmidrule[0.5pt]{9-15}
Data sets & CC & LL & RPC && IB-M & IB-PL && LRT & \textbf{LR-RF} & LRT+bag & W-LRT+bag & F-LRT+bag & LRF+RT & LRT+ERT\\
\midrule[0.5pt]
authorship &0.920(3.0) &0.657(12.0) &0.910(9.0) & &\textbf{0.936}(1.5) &\textbf{0.936}(1.5) & &0.882(11.0) &0.913(7.0)&0.916(4.5)&0.914(6.0)&0.916(4.5) &0.912(8.0) & 0.906(10.0)\\
bodyfat    &0.281(2.0) &0.266(3.0) &\textbf{0.285}(1.0) & &0.229(6.0) &0.230(5.0) & &0.117(12.0) &0.185(11.0)&0.233(4.0)&0.225(7.0)&0.210(10.0) & 0.212(8.0) & 0.211(9.0)\\
calhousing &0.250(9.0) &0.223(11.0) &0.243(10.0) & &0.344(5.0) &0.326(6.0) & &0.324(7.0) &0.367(4.0)&\textbf{0.476}(1.0)&0.447(2.0)&0.429(3.0)& 0.185(12.0) & 0.294(8.0)\\
cup-small  &0.475(7.0) &0.419(12.0) &0.450(10.0) & &0.496(5.0) &0.495(6.0) & &0.447(11.0) &0.515(2.5)&\textbf{0.524}(1.0)&0.498(4.0)&0.515(2.5)&0.469(9.0) &0.471(8.0)\\
elevators  &0.768(4.0) &0.701(11.0) &0.749(7.0) & &0.727(8.0) &0.721(9.5) & &0.760(5.0) &0.756(6.0)&\textbf{0.790}(1.0)&0.784(3.0)&0.785(2.0)&0.605(12.0)&0.721(9.5)\\
fried      &\textbf{0.999}(1.5) &0.989(3.0) &\textbf{0.999}(1.5) & &0.900(7.0) &0.894(8.0) & &0.890(10.0) &0.926(6.0)&0.955(4.0)&0.891(9.0)&0.929(5.0)&0.887(11.0)&0.841(12.0)\\
glass      &0.846(7.0) &0.818(11.0) &0.882(3.0) & &0.842(8.0) &0.841(9.0) & &0.883(2.0) &\textbf{0.888}(1.0)&0.856(5.0)&0.813(12.0)&0.839(10.0)&0.874(4.0)&0.849(6.0)\\
housing    &0.660(11.0) &0.626(12.0) &0.671(10.0) & &0.736(6.0) &0.711(8.0) & &\textbf{0.797}(1.0) &0.792(2.0)&0.761(4.0)&0.715(7.0)&0.738(5.0)&0.780(3.0)&0.699(9.0)\\
iris       &0.836(11.0) &0.818(12.0) &0.889(10.0) & &0.925(9.0) &0.960(4.0) & &0.947(7.0) &0.966(2.0)&0.956(5.5)&0.961(3.0)&0.956(5.5)&\textbf{0.973(1.0)}&0.933(8.0)\\
pendigits  &0.903 &0.814 &0.932 & &0.941 &0.939 & &0.935 &0.939)&\textbf{0.945}&0.925&0.929&*&*\\
segment    &0.914(10.0) &0.810(11.0) &0.934(7.0) & &0.802(12.0) &0.950(3.0) & &0.949(4.0) &\textbf{0.961}(1.0)&0.958(2.0)&0.944(6.0)&0.946(5.0)&0.930(8.0)&0.917(9.0)\\
stock      &0.737(11.0) &0.696(12.0) &0.777(10.0) & &\textbf{0.925}(1.0) &0.922(2.5) & &0.895(6.0) &0.922(2.5)&0.907(4.0)&0.894(7.0)&0.901(5.0)&0.892(8.0)&0.869(9.0)\\
vehicle    &0.855(5.5) &0.770(12.0) &0.854(7.0) & &0.855(5.5) &0.859(3.0) & &0.827(11.0) &0.860(2.0)&\textbf{0.867}(1.0)&0.858(4.0)&0.850(8.5)&0.850(8.5)&0.849(10.0)\\
vowel      &0.623(11.0) &0.601(12.0) &0.647(10.0) & &\textbf{0.882}(1.0) &0.851(3.0) & &0.794(5.0) &0.867(2.0)&0.758(6.0)&0.711(8.0)&0.719(7.0)&0.844(4.0)&0.701(9.0)\\
wine       &0.933(5.0) &0.942(4.0) &0.921(10.0) & &0.944(3.0) &0.947(2.0) & &0.882(12.0) &\textbf{0.953}(1.0)&0.924(9.0)&0.928(7.0)&0.916(11.0)&0.932(6.0)&0.925(8.0)\\
wisconsin  &0.629(2.0) &0.542(3.0) &\textbf{0.633}(1.0) & &0.501(4.0) &0.479(5.0) & &0.343(10.0) &0.478(6.0)&0.389(9.0)&0.340(12.0)&0.384(11.0)&0.460(7.0)&0.429(8.0)\\
\midrule[0.5pt]
avg.rank   & 6.67     & 9.40      & 7.10      & & 5.47      & 5.03      & & 7.60      & \textbf{3.73} & 4.07 & 6.47 & 6.33  &7.30 & 8.83 \\
\bottomrule[0.75pt]
\end{tabular}
\end{tiny}
\end{center}
\end{sidewaystable}

\begin{sidewaystable}
\caption{Comparative results between LR-RF with state-of-the-art algorithms on the case $p_0 = 0.30$.}
\label{Tab:Comparison Results P = 0.3}
\begin{center}
\begin{scriptsize}
\begin{tabular}{l|cccccccccccc}
\toprule[0.75pt]
\multicolumn{1}{c}{} & \multicolumn{3}{c}{Reduction methods} & \multicolumn{1}{c}{} & \multicolumn{2}{c}{Probabilistic methods} & \multicolumn{1}{c}{} & \multicolumn{5}{c}{Tree-based methods}\\
\cmidrule[0.5pt]{2-4} \cmidrule[0.5pt]{6-7} \cmidrule[0.5pt]{9-13}
Data sets & CC & LL & RPC && IB-M & IB-PL && LRT & \textbf{LR-RF} & LRT+bag & W-LRT+bag & F-LRT+bag\\
\midrule[0.5pt]
authorship &0.891(7.0) &0.656(10.0) &0.884(8.0) & &0.913(3.0) &\textbf{0.927}(1.0) & &0.871(9.0) &0.911(5.0)&0.912(4.0)&0.907(6.0)&0.916(2.0)\\
bodyfat    &0.260(2.0) &0.251(3.0) &\textbf{0.272}(1.0) & &0.198(8.0) &0.204(6.0) & &0.097(10.0) &0.181(9.0)&0.218(4.0)&0.209(5.0)&0.200(7.0)\\
calhousing &0.249(8.0) &0.223(10.0) &0.243(9.0) & &0.310(5.0) &0.303(7.0) & &0.307(6.0) &0.364(4.0)&\textbf{0.457}(1.0)&0.435(2.0)&0.407(3.0)\\
cup-small  &0.474(6.0) &0.419(9.0) &0.449(8.0) & &0.473(7.0) &0.477(5.0) & &0.405(10.0) &0.513(2.0)&\textbf{0.517}(1.0)&0.490(4.0)&0.505(3.0)\\
elevators  &0.767(4.0) &0.699(9.0) &0.748(10.0) & &0.683(7.0) &0.702(8.0) & &0.756(5.0) &0.755(6.0)&\textbf{0.789}(1.0)&0.780(3.0)&0.782(2.0)\\
fried      &0.998(2.0) &0.989(3.0) &\textbf{0.999}(1.0) & &0.850(10.0) &0.861(9.0) & &0.863(8.0) &0.924(5.0)&0.949(4.0)&0.877(7.0)&0.915(6.0)\\
glass      &0.835(5.0) &0.817(7.0) &0.851(2.0) & &0.776(10.0) &0.809(8.0) & &0.850(3.0) &\textbf{0.886}(1.0)&0.845(4.0)&0.801(9.0)&0.828(6.0)\\
housing    &0.655(8.0) &0.625(10.0) &0.667(7.0) & &0.669(9.0) &0.654(6.0) & &0.734(3.0) &\textbf{0.789}(1.0)&0.744(2.0)&0.702(5.0)&0.729(4.0)\\
iris       &0.807(9.0) &0.804(10.0) &0.871(7.0) & &0.867(8.0) &0.926(5.0) & &0.909(6.0) &\textbf{0.962}(1.0)&0.941(4.0)&0.943(3.0)&0.945(2.0)\\
pendigits  &0.902(8.5) &0.802(10.0) &0.932(3.0) & &0.902(8.5) &0.918(6.0) & &0.914(7.0) &0.939(2.0)&\textbf{0.945}(1.0)&0.920(5.0)&0.924(4.0)\\
segment    &0.911(7.0) &0.806(9.0) &0.933(5.5) & &0.735(10.0) &0.874(8.0) & &0.933(5.5) &\textbf{0.961}(1.0)&0.953(2.0)&0.940(4.0)&0.942(3.0)\\
stock      &0.735(9.0) &0.691(10.0) &0.776(8.0) & &0.855(7.0) &0.877(5.5) & &0.877(5.5) &\textbf{0.922}(1.0)&0.897(2.0)&0.888(4.0)&0.889(3.0)\\
vehicle    &0.839(5.0) &0.769(10.0) &0.834(7.0) & &0.822(8.0) &0.838(6.0) & &0.819(9.0) &0.859(2.0)&\textbf{0.866}(1.0)&0.852(3.0)&0.847(4.0)\\
vowel      &0.615(9.0) &0.598(10.0) &0.644(8.0) & &0.810(2.0) &0.785(3.0) & &0.718(5.0) &\textbf{0.851}(1.0)&0.746(4.0)&0.708(7.0)&0.714(6.0)\\
wine       &0.911(8.0) &0.941(2.0) &0.902(9.0) & &0.930(3.0) &0.926(4.0) & &0.862(10.0) &\textbf{0.952}(1.0)&0.923(5.0)&0.915(7.0)&0.918(6.0)\\
wisconsin  &\textbf{0.617}(1.0) &0.533(3.0) &0.607(2.0) & &0.464(5.0) &0.453(6.0) & &0.284(10.0) &0.468(4.0)&0.388(7.0)&0.331(9.0)&0.374(8.0)\\
\midrule[0.5pt]
avg.rank   & 6.16      & 7.81      & 5.97      & & 6.91      &5.84       & & 7.00      & \textbf{2.88} & 2.94 & 5.19 & 4.31 \\
\bottomrule[0.75pt]
\end{tabular}
\end{scriptsize}
\end{center}
\end{sidewaystable}

\begin{sidewaystable}
\caption{Comparative results between LR-RF with state-of-the-art algorithms on the case $p_0 = 0.60$.}
\label{Tab:Comparison Results P = 0.6}
\begin{center}
\begin{scriptsize}
\begin{tabular}{l|cccccccccccc}
\toprule[0.75pt]
\multicolumn{1}{c}{} & \multicolumn{3}{c}{Reduction methods} & \multicolumn{1}{c}{} & \multicolumn{2}{c}{Probabilistic methods} & \multicolumn{1}{c}{} & \multicolumn{5}{c}{Tree-based methods}\\
\cmidrule[0.5pt]{2-4} \cmidrule[0.5pt]{6-7} \cmidrule[0.5pt]{9-13}
Data sets & CC & LL & RPC && IB-M & IB-PL && LRT & \textbf{LR-RF}  & LRT+bag & W-LRT+bag & F-LRT+bag\\
\midrule[0.5pt]
authorship &0.835(8.0) &0.650(10.0) &0.872(6.0) & &0.849(7.0) &0.886(5.0) & &0.828(9.0) &0.897(3.0)&0.905(2.0)&0.895(4.0)&\textbf{0.907}(1.0)\\
bodyfat &0.224(3.0) &\textbf{0.241}(1.0) &0.235(2.0) & &0.160(6.0) &0.151(8.0) & &0.007(10.0) &0.095(9.0)&0.180(4.0)&0.175(5.0)&0.156(7.0)\\
calhousing &0.247(8.0) &0.221(10.0) &0.242(9.0) & &0.263(6.0) &0.259(7.0) & &0.273(5.0) &0.363(2.0)&\textbf{0.400}(1.0)&0.356(3.0)&0.330(4.0)\\
cup-small  &0.470(5.0) &0.418(9.0) &0.448(6.0) & &0.429(8.0) &0.437(7.0) & &0.367(10.0) &\textbf{0.511}(1.0)&0.499(2.0)&0.471(4.0)&0.488(3.0)\\
elevators  &0.765(4.0) &0.696(8.0) &0.748(6.0) & &0.596(10.0) &0.633(9.0) & &0.742(7.0) &0.755(5.0)&\textbf{0.782}(1.0)&0.767(3.0)&0.771(2.0)\\
fried      &\textbf{0.997}(1.5) &0.987(3.0) &\textbf{0.997}(1.5) & &0.777(10.0) &0.797(9.0) & &0.809(8.0) &0.921(5.0)&0.932(4.0)&0.845(7.0)&0.883(6.0)\\
glass      &0.789(7.0) &0.808(3.0) &0.799(5.5) & &0.611(10.0) &0.675(9.0) & &0.799(5.5) &\textbf{0.867}(1.0)&0.825(2.0)&0.781(8.0)&0.801(4.0)\\
housing    &0.638(6.0) &0.614(8.0) &0.641(5.0) & &0.543(9.0) &0.492(10.0) & &0.634(7.0) &\textbf{0.776}(1.0)&0.699(2.0)&0.672(4.0)&0.690(3.0)\\
iris       &0.743(10.0) &0.768(8.0) &0.779(7.0) & &0.799(5.0) &0.868(2.0) & &0.794(6.0) &\textbf{0.959}(1.0)&0.852(3.0)&0.765(9.0)&0.828(4.0)\\
pendigits  &0.900(6.0) &0.787(9.0) &0.929(3.0) & &0.781(10.0) &0.794(8.0) & &0.871(7.0) &\textbf{0.938}(1.0)&0.935(2.0)&0.907(5.0)&0.912(4.0)\\
segment    &0.902(7.0) &0.801(8.0) &0.920(5.0) & &0.612(10.0) &0.674(9.0) & &0.903(6.0) &\textbf{0.957}(1.0)&0.940(2.0)&0.923(4.0)&0.932(3.0)\\
stock      &0.724(8.5) &0.689(10.0) &0.771(6.0) & &0.724(8.5) &0.740(7.0) & &0.827(5.0) &\textbf{0.912}(1.0)&0.870(2.0)&0.855(4.0)&0.857(3.0)\\
vehicle    &0.810(5.0) &0.764(7.5) &0.786(6.0) & &0.736(10.0) &0.765(9.0) & &0.764(7.5) &\textbf{0.854}(1.0)&0.841(2.0)&0.818(4.0)&0.820(3.0)\\
vowel      &0.598(8.0) &0.591(9.0) &0.612(7.0) & &0.638(5.0) &0.588(10.0) & &0.615(6.0) &\textbf{0.787}(1.0)&0.722(2.0)&0.694(4.0)&0.697(3.0)\\
wine       &0.853(7.0) &0.894(3.0) &0.864(6.0) & &0.893(4.0) &0.907(2.0) & &0.752(8.0) &\textbf{0.926}(1.0)&0.867(5.0)&0.809(9.0)&0.802(10.0)\\
wisconsin  &\textbf{0.566}(1.0) &0.518(3.0) &0.536(2.0) & &0.399(4.0) &0.381(5.0) & &0.251(10.0) &0.371(7.0)&0.372(6.0)&0.310(9.0)&0.356(8.0)\\
\midrule[0.5pt]
avg.rank   & 5.94      & 6.84      & 5.19      & & 7.66      & 7.25      & & 7.31      & \textbf{2.56} & 2.63 & 5.38 & 4.25     \\
\bottomrule[0.75pt]
\end{tabular}
\end{scriptsize}
\end{center}
\end{sidewaystable}

To compare our LR-RF algorithm with these reference algorithms, we use mentioned-above two-step statistical test procedure, i.e., a Friedman test and a two-tailed Bonferroni-Dunn test. According to the average ranks of each methods, we can calculate $F_F = 4.18, 6.67$ and $9.34$ for cases with complete ranking, $30\%$ missing labels and $60\%$ missing labels, respectively. All the three $F_F$ values are larger than the critical value of the Friedman test $F_{0.05}(11,154) = 1.79$ and $F_{0.05}(9,135) = 1.92$. Consequently, the Friedman test rejects the null hypothesis in all three cases, which suggests these seven methods are not equivalent in all three cases. After rejecting the null-hypothesis, we conduct two-tailed Bonferroni-Dunn test for each case.

Table \ref{Tab:Comparison Results P = 0.0} shows the comparative results in case of complete ranking information. At the significance level 0.05, CD is $2.871 \sqrt{\frac{12 \cdot 13}{6 \cdot 15}} = 3.78$. From Table \ref{Tab:Comparison Results P = 0.0}, we only observe that our proposed LR-RF algorithm significantly outperforms LL ($9.40-3.73 = 5.67 > 3.78$), LRT ($7.60-3.73 = 3.87 > 3.78$) and LRT+ERT ($8.83-3.73 = 5.10 > 3.78$). Compared to other reference algorithms, our LR-RF achieves better performance in terms of average rank but there is no significant differences between their performance.

Table \ref{Tab:Comparison Results P = 0.3} summarizes the comparative results in case of incomplete ranking information with $p = 0.3$. At the significance level 0.05, CD is $2.773 \sqrt{\frac{10 \cdot 11}{6 \cdot 16}} = 2.97$. As we can see from Table \ref{Tab:Comparison Results P = 0.3}, our proposed LR-RF also achieves the best performance in terms of average rank, and it significantly outperforms reference algorithms CC ($6.16-2.88 = 3.28 > 2.97$), LL ($7.81-2.88 = 4.93$), RPC ($5.97-2.88 = 3.09 > 2.97$), IB-M ($6.91-2.88 = 4.03 > 2.97$) and LRT ($7.00-2.88 = 4.12 > 2.97$). We also observe that our LR-RF is better than IB-PL algorithm, and the critical value is only slightly less than CD value, i.e., $5.84-2.88 = 2.96 < 2.97$.

Table \ref{Tab:Comparison Results P = 0.6} presents the comparative results in case of incomplete ranking information with $p = 0.6$. At the significance level 0.05, CD is also 2.97. From this table, we observe that our proposed LR-RF algorithm significantly outperforms reference algorithms CC ($5.94-2.56 = 3.38 > 2.97$), LL ($6.84-2.56 = 4.28 > 2.97$), IB-M ($7.66 - 2.56 = 5.10 > 2.97$), IB-PL ($7.25-2.56 = 4.69 > 2.97$) and LRT ($7.31-2.56 = 4.75 > 2.97$). Compared to reference algorithm W-LRT+bag, the difference is slightly less than corresponding CD (i.e., $5.38 - 2.56 = 2.82 < 2.97$).

Besides the above observations, we also can draw from Table \ref{Tab:Comparison Results P = 0.0}-\ref{Tab:Comparison Results P = 0.6}:
\begin{itemize}
	\item Comparing performances of LR-RF in these three cases, we find that our LR-RF has stronger ability to deal with data set with partial ranking information. When the probability of missing labels $p_0$ becomes higher, the advantage of the LR-RF is more obvious. When $p_0 = 0$ increase to $p_0 = 0.3$ and $p_0 = 0.6$, we find the average ranks of LR-RF decrease and achieves the best performance on much more data sets.
	\item Comparing with four tree-based methods, our LR-RF achieves the best performance in terms of average rank on all three cases. Our proposed LR-RF significantly outperforms LRT in all cases. This results confirm the conclusion that a ensemble model (e.g., LR-RF) can produce substantial improvements in the performance compared with the use of a single model (e.g., LRT).
	\item Comparing with reduction methods and probabilistic methods, our LR-RF also shows highly competitive performance. In case of complete ranking information, our LR-RF significant outperforms LL. In case of incomplete ranking information, our LR-RF demonstrates better performance, and is significantly outperforms algorithms CC, LL, IB-M and even RPC and IB-PL. This results show that our LR-RF is highly competitive compared with the state-of-the-art algorithms.
\end{itemize}

\section{Discussion and analysis}
\label{Sec:Discussion and Analysis}

This section is devoted to an experimental analysis and discussion of the proposed LR-RF algorithm. Specifically, we investigate the impact of the noise on the performance of the proposed LR-RF algorithm, and conduct sensitive analysis on two important parameters, i.e., $nbr_{tree}$ and $d_{max}$. These three experimental analyses are based on four representative data sets (i.e., glass, housing, iris, and wine) selected from benchmarks.

\subsection{The impact of noise on the performance of our LR-RF algorithm.}
\label{SubSec:The Impact of Noise on the Performance of our LR-RF algorithm}

\begin{figure}[!htb]
\begin{center}
\centerline{\includegraphics[width=0.9\textwidth]{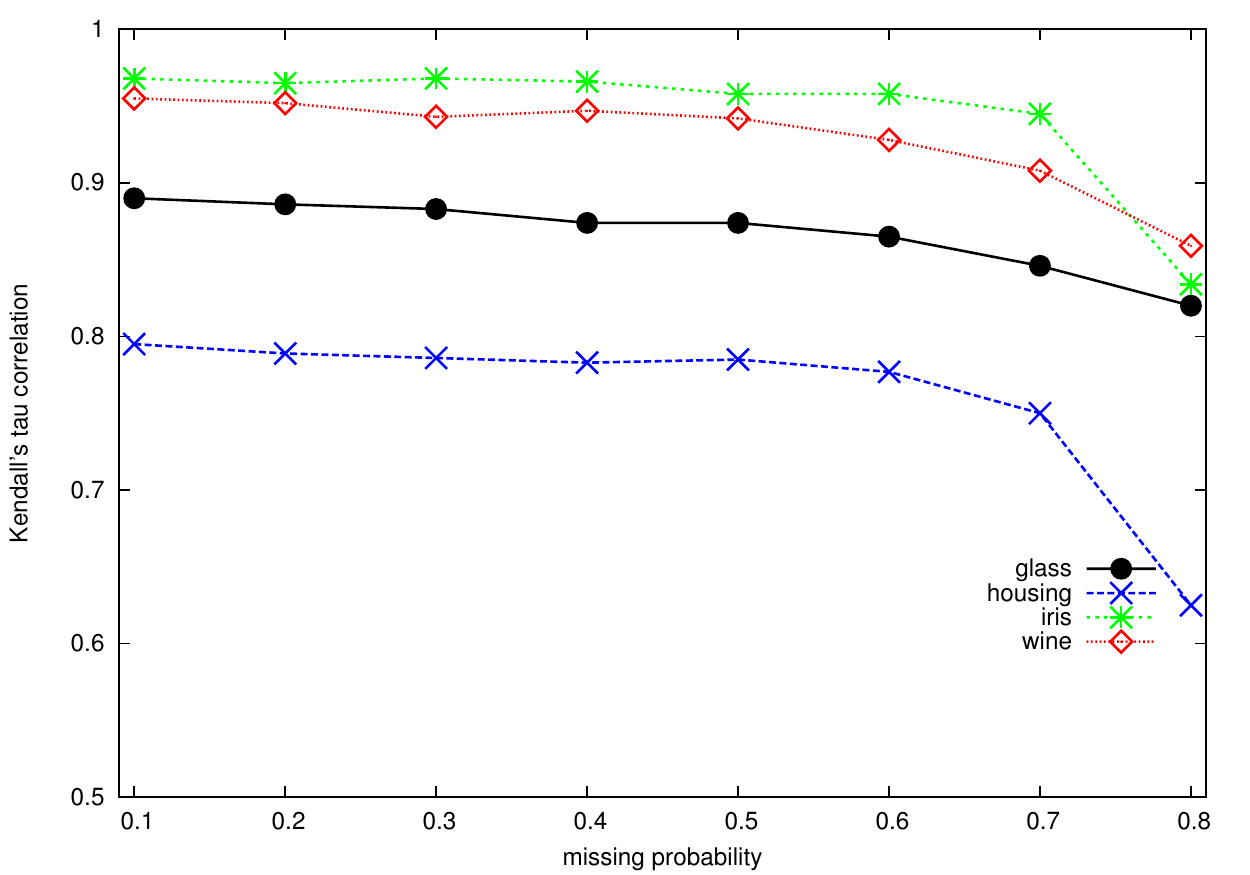}}
\caption{Ranking performance (in terms of Kendall's tau correlation) of the LR-RF as a function of the probability of missing label $p_0$.}
\label{Fig:Learning Curves}
\end{center}
\end{figure}

In this set of experiments, we perform experiments to gain some understanding about how the noise at the label information (i.e., the missing probability $p_0$) affects the performance of the proposed LR-RF algorithm. We did experiments with LR-RF on four selected data sets for different missing probability $p_0$, varying $p_0$ from $0.1$ to $0.8$ by steps of $0.1$. Figure \ref{Fig:Learning Curves} shows the learning curves, which present the performance of LR-RF as a function of the missing probability $p_0$.

From Figure \ref{Fig:Learning Curves}, we can clearly observe that the learning curves are relatively flat when $p_0$ varies from 0.10 to 0.60. When the probability of missing label $p_0$ continues to increase, the performance of our LR-RF algorithm rapidly worsening. The results indicate that our proposed method LR-RF has a strong robustness, and it is able to deal with data with missing label ranking information. It agrees with the conclusions obtained from the computational studies in Section \ref{SubSec:Compared With Reported Results of the State-of-the-art Algorithms}.

\subsection{Sensitivity analysis to the $nbr_{tree}$ parameter}
\label{SubSec:Sensitivity Analysis on the Number of Trees}

The third set of experiments investigate how the parameter $nbr_{tree}$ affects the accuracy of the proposed LR-RF method. We did experiments on data with complete ranking for different $nbr_{tree}$ values, varying $nbr_{tree}$ from 1 to 75 by steps of 5, and fixing other parameter value at the same time. Specifically, when $nbr_{tree}$ equals to 1, the LR-RF is a single decision tree. This label ranking tree distinguishes itself from LRT \cite{Cheng2009} by the way to use ranking information for guiding the construction of a decision tree.

\begin{figure}[!ht]
\begin{center}
\centerline{\includegraphics[width=0.9\textwidth]{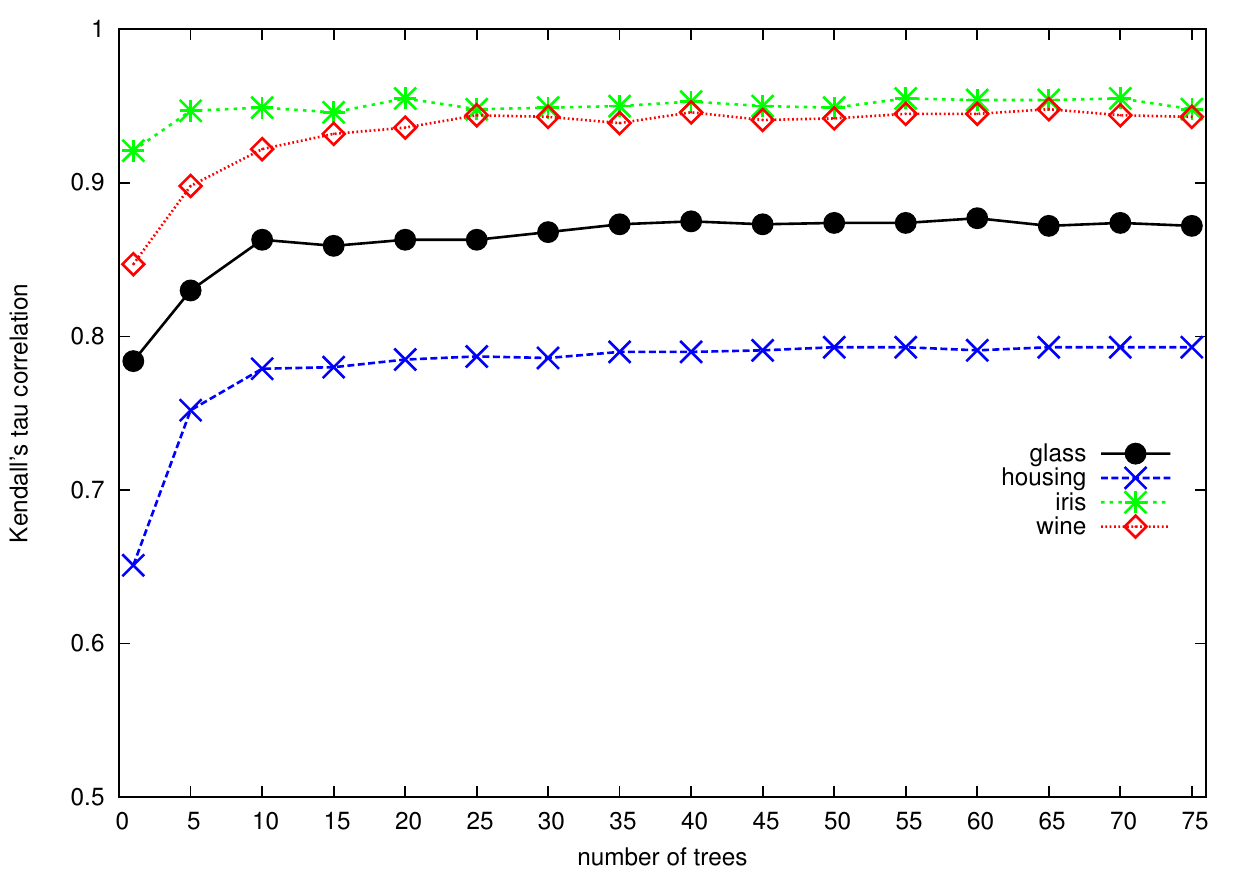}}
\caption{Ranking performance (in terms of Kendall's tau correlation) of the LR-RF as a function of the number of trees $nbr_{tree}$.}
\label{Fig:Sensitivity Analysis on the Number of Trees}
\end{center}
\end{figure}

In Figure \ref{Fig:Sensitivity Analysis on the Number of Trees}, we are able to see how the different number of trees $nbr_{tree}$ affects the performance of the LR-RF. It is interesting to find that, when $nbr_{tree}$ increases from $1$ to $10$, the computational accuracy of the LR-RF algorithm significantly improves, while the computational accuracy of the LR-RF algorithm keeps relatively steady if $nbr_{tree}$ continues to increase until $75$. We can draw a preliminary conclusion that an ensemble (e.g., random forest) of decision trees can outperforms a single decision tree. In addition, this analysis also shows that a reasonable $nbr_{tree}$ value plays an important role in the success of the LR-RF. Fortunately, the reasonable value of parameter $nbr_{tree}$ has a large range. It is easy to determine a suitable value for parameter $nbr_{tree}$. In our proposed method LR-RF, we roughly fix the total number of decision tree $nbr_{tree}$ as $50$.

\subsection{Sensitivity analysis to the $d_{max}$ parameter}
\label{SubSec:Sensitivity Analysis on Tree Depth}

We also study how the tree depth $d_{max}$ affects the performance of the proposed LR-RF algorithm. We did experiments on data with complete ranking for different $d_{max}$ values, varying $d_{max}$ from 1 to 15 by steps of 1, and fixing other parameter value at the same time.

\begin{figure}[!ht]
\begin{center}
\centerline{\includegraphics[width=0.9\textwidth]{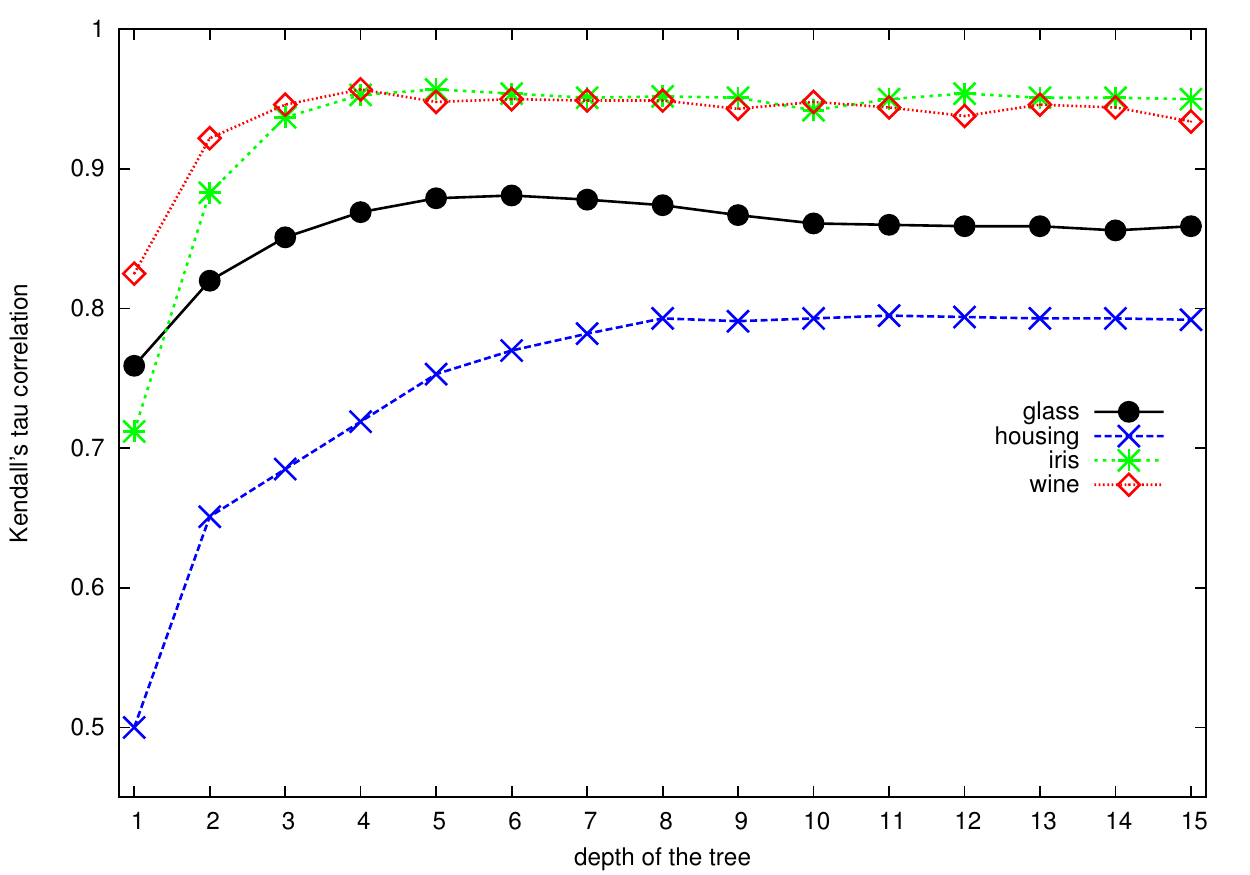}}
\caption{Ranking performance (in terms of Kendall's tau correlation) of the LR-RF as a function of the depth of tree $d_{max}$.}
\label{Fig:Sensitivity Analysis on the Tree Depth}
\end{center}
\end{figure}

Figure \ref{Fig:Sensitivity Analysis on the Tree Depth} describes the influence of tree depth $d_{max}$ on the performance of the proposed LR-RF algorithm. When the value of parameter $d_{max}$ is small, we observe that as the $d_{max}$ increases the overall performance of LR-RF algorithm also increases, and then the performance of the LR-RF algorithm keeps a relative steady value if we continue to increase the value of parameter $d_{max}$. We clearly observe that the performance of LR-RF algorithms keeps steady in a wide range, i.e., $d_{max} \geqslant 8$. Moreover, the tree depth is a function of the problem complexity. For a given $d_{max}$ value, the number of possible nodes is up to $2^{d_{max}}$ in a random tree. Therefore, in our algorithm, we roughly set the value of parameter $d_{max}$ as 8.

\section{Conclusions and future work}
\label{Sec:Conclusions and Future Work}

In this paper, we have developed a novel label ranking method based on random forest (LR-RF). The empirical results show that our method is highly competitive with state-of-the-art methods in terms of predictive accuracy for datasets with complete ranking and datasets with partial ranking information. In addition to achieving state of the art performances, the new method has some further advantages. Firstly, the tree structure of our LR-RF makes the retrieval of nearest neighbours more efficient and scalable than traditional instance-based approaches. Secondly, both the construction and prediction processes of our method can be executed in a parallel way. Thirdly, our LR-RF method is a tree-based method, tree structure can clearly express much more information about the problem and it is easy to understand even for people without a background on learning algorithms. It is worth noting that our proposed LR-RF shares the third and fouth advantage with two recent algorithms proposed in \cite{Aledo2017} and \cite{DeSa2017}.

We also performed three sets of experiments to investigate the impact of noise on the performance of the LR-RF, and study the influence of parameter $nbr_{tree}$ and $d_{max}$ on the accuracy of the LR-RF, respectively. The analysis results also show that our method has a strong robustness on the missing label information and parameter values selection.

For future work, we plan to compare and analyse the different methods to use ranking information for the construction of a decision tree, and select the suitable one to guide the growing of the decision trees in a random forest. It is possible to further improve the performance of LR-RF by use the suitable method for using the ranking information.

\section*{Acknowledgment}

We are grateful to our reviewers for their useful comments and suggestions which helped us to improve the paper.

\bibliography{mybibfiles}
\end{document}